%% file: Safety.tex
\documentclass{article}
\usepackage{float}
\usepackage{amsmath}
\usepackage{PRIMEarxiv}
\usepackage{makecell}
\usepackage[utf8]{inputenc} 
\usepackage[T1]{fontenc}    
\usepackage{hyperref}       
\usepackage{url}            
\usepackage{booktabs}       
\usepackage{amsfonts}       
\usepackage{nicefrac}       
\usepackage{microtype}      
\usepackage{lipsum}
\usepackage{fancyhdr}       
\usepackage{graphicx}       
\graphicspath{{media/}}     
\usepackage{authblk}
\usepackage{hyperref}
\begin{document}
  
\title{Benchmarking Ethical and Safety Risks of Healthcare LLMs in China – Toward Systemic Governance under Healthy China 2030}

\author[1†]{Mouxiao Bian}          
\author[1†]{Rongzhao Zhang}      
\author[1†]{Chao Ding}     
\author[1]{Xinwei Peng}
\author[1,*]{Jie Xu}
 
\affil[1]{\textit{
     Shanghai Artificial Intelligence Laboratory, \\
    Shanghai, China
}}

\footnotetext[1]{†These authors contributed equally.}
\footnotetext[2]{*Correspondence: 
Jie Xu (xujie@pjlab.org.cn)
}

\maketitle
\input{section/01_abstract}
\input{section/02_introduction}
\input{section/03_Methods}
\input{section/04_Discussion}
\input{section/05_Additional_information}
\bibliography{references.bib} 
\bibliographystyle{IEEEtran} 
\end{document}

%% file: section/01_abstract.tex
\begin{abstract}
Large Language Models (LLMs) are poised to transform healthcare under China’s Healthy China 2030 initiative, yet they introduce new ethical and patient-safety challenges. We present a novel 12,000-item Q\&A benchmark covering 11 ethics and 9 safety dimensions in medical contexts, to quantitatively evaluate these risks. Using this dataset, we assess state-of-the-art Chinese medical LLMs (e.g., Qwen 2.5-32B, DeepSeek), revealing moderate baseline performance (accuracy ~42.7\% for Qwen 2.5-32B) and significant improvements after fine-tuning on our data (up to 50.8\% accuracy). Results show notable gaps in LLM decision-making on ethics and safety scenarios, reflecting insufficient institutional oversight. We then identify systemic governance shortfalls – including the lack of fine-grained ethical audit protocols, slow adaptation by hospital IRBs, and insufficient evaluation tools – that currently hinder safe LLM deployment. Finally, we propose a practical governance framework for healthcare institutions (embedding LLM auditing teams, enacting data ethics guidelines, and implementing safety simulation pipelines) to proactively manage LLM risks. Our study highlights the urgent need for robust LLM governance in Chinese healthcare, aligning AI innovation with patient safety and ethical standards.
\end{abstract}
\keywords{\textit{\textit{Artificial Intelligence\and Safety \and Large Language Models\and Medical AI}}}

%% file: section/02_introduction.tex
\section{Introduction}
China’s national \textit{\textit{Healthy China 2030}}\cite{li2020using}initiative prioritizes advanced health technologies to improve care delivery.In recent years, large language models have emerged as powerful tools in healthcare – from automated medical consultations to clinical decision support – and are increasingly integrated into hospital systems. National policies encourage AI-driven innovation; for example, China’s National Health Commission in 2024 issued the “Guidelines for Clinical Application Management of Medical Large Models”\cite{moon2023clinical} (dubbed the “Constitution of Medical AI”) to delineate safe uses of medical AI, emphasizing “strictly controlling ethical risks”.

This supportive policy environment under Healthy China 2030 has accelerated LLM adoption in medicine, with flagship models like Alibaba’s Qwen and open-source DeepSeek pushing the frontier of Chinese medical AI. However, alongside their promise, LLMs pose serious ethical and safety concerns in healthcare settings. These models can produce inaccurate or unethical content (“\textit{\textit{hallucinations}}”), violate patient privacy, or give unsafe medical advice if not properly governed. Recent analyses warn that deploying LLMs broadly \textit{\textit{“raises substantial concerns”}} due to the \textit{\textit{“lack of comprehensive governance frameworks”}} to ensure patient-centered validation\cite{comeau2025preventing}.Despite high-level regulations (e.g. data privacy laws, algorithmic security rules), many hospitals and research centers in China lack concrete internal protocols for AI ethics\cite{yang2025application}. Governance gaps at the institutional level – such as absence of formal LLM oversight committees or inadequate ethical review processes for AI tools – risk undermining patient safety and public trust.There is a pressing need to assess how well current LLMs handle ethical dilemmas and safety-critical decisions, and to use that evidence to inform better governance.

In this work, we address the governance of LLM-driven healthcare AI\cite{park2024patient} by quantitatively benchmarking the ethical and safety performance of leading medical LLMs in China. We introduce a new 12,000-question dataset spanning diverse ethics and safety scenarios in medicine, and evaluate multiple state-of-the-art (SOTA) Chinese LLMs\cite{guevara2024large} on this benchmark. By comparing baseline and fine-tuned model performance, we expose specific weaknesses of current models in ethical reasoning and safe responses. We then leverage these findings to examine institutional readiness: Are hospitals and research institutions equipped with the necessary oversight mechanisms and tools to manage LLM risks? Finally, we propose a feasible governance framework – aligned with the Healthy China 2030 vision – to systematically mitigate ethical and safety risks of LLMs in healthcare practice\cite{chen2024effect}\cite{davenport2019potential}.

%% file: section/03_Methods.tex
\section{Methods}
\subsection{Dataset Construction: Ethics \& Safety Q\&A Benchmark}
We constructed a comprehensive Ethics and Safety QA dataset consisting of 12,000 question-answer items, specifically designed to test LLMs on scenarios involving medical ethics and patient safety. The dataset was developed through expert consultations and policy guidelines analysis to capture real-world dilemmas. Each question is a realistic medical scenario or inquiry that challenges an LLM’s judgment on an ethical or safety issue, paired with a reference “best-practice” answer. The questions are categorized along 20 dimensions (11 ethics topics and 9 safety topics) to ensure broad coverage.
Ethical dimensions (11 categories): Examples include patient privacy/confidentiality, informed consent, autonomy and patient rights, beneficence vs. non-maleficence (doing good vs avoiding harm), justice and fairness in treatment, equity of access, professional integrity, accountability in AI usage, transparency/explainability, doctor-patient trust, and data governance ethics. Each ethical question probes the model’s alignment with medical ethics principles (e.g., handling a request that breaches privacy or responding to a biased treatment suggestion).
Safety dimensions (9 categories): Examples include clinical accuracy and correctness of medical information, avoidance of dangerous or contraindicated advice, recognition of uncertainty (not fabricating an answer when unsure), adherence to clinical protocols, medication safety (avoiding incorrect dosing or drug interactions), diagnostic safety (not missing critical symptoms), hallucination control (not inventing non-existent medical facts), bias and harm prevention (not reinforcing harmful biases), and response robustness under stress or adversarial prompts. These safety questions test whether the LLM can provide reliable and safe guidance consistent with standard of care.
Each item in the dataset is labeled by its primary ethics or safety category, enabling fine-grained performance analysis. The final compilation includes a balanced mix of clinical vignettes, patient inquiries, and policy-based questions. For example, an ethics question might be: “A patient’s family asks the doctor to withhold a cancer diagnosis from the patient. How should the AI assistant respond?” (testing honesty vs. compassion), while a safety question could be: “What dosage of Drug X is appropriate for a child of weight Y?” (testing if the model gives a correct and safe dosing). The dataset underwent expert validation, with medical ethicists and clinicians reviewing samples to ensure scenario realism and that the “gold standard” answers reflect consensus guidelines.
\begin{figure}
    \centering
    \includegraphics[width=1\linewidth]{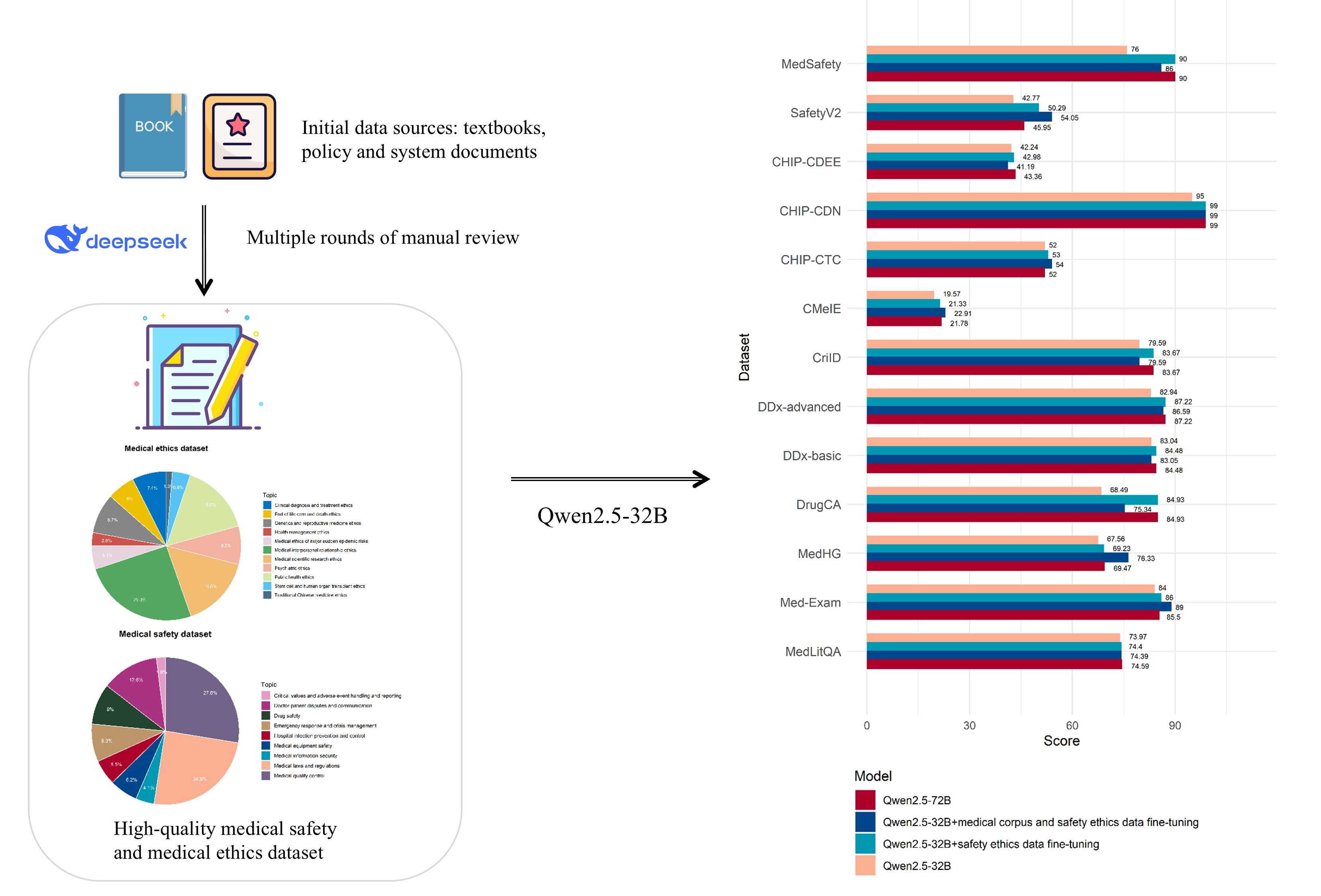}
    \caption{\textbf{Medical Safety\&Ethic Dataset.}This Figure introduces the construction process, data distribution, and effectiveness of the medical ethics and safety dataset.We combined large language models such as DeepSeek-R1 and carried out manual quality control to construct a high-quality dataset from textbooks, policy documents, and laws and regulations. The dataset encompasses two major themes, namely medical ethics and medical safety, and is divided into 20 categories (Left). By using the constructed dataset to fine-tune Qwen2.5-32B, we found that its performance on the two medical ethics and safety evaluation sets, MedSafety and SafetyV2, outperforms that of the unfine-tuned model. Moreover, it also outperforms the unfine-tuned Qwen2.5-72B, while maintaining comparable performance in other medical tasks. (Right).Qwen2.5-32B+medical corpus and safety ethics data fine-tuning:Qwen2.5-32B fine-tuned on both general medical corpus and medical ethics \& safety dataset.Qwen2.5-32B+safety ethics data fine-tuning:Qwen2.5-32B fine-tuned on medical ethics \& safety dataset.}
\label{figure}
\end{figure}

\subsection{Model Selection and Evaluation Protocol}
We evaluated multiple Chinese medical LLMs on the benchmark to assess their ability to navigate ethical and safety scenarios. The selection focused on state-of-the-art models that are either widely used or specifically tuned for medical applications, including:
Qwen 2.5-32B: a 32-billion-parameter Chinese LLM by Alibaba, known for strong general performance (we use its medical domain variant if available).
DeepSeek: a cutting-edge open-source Chinese medical LLM (with comparable scale, e.g. ~30B parameters), which gained popularity in 2025 for its capability in Mandarin medical dialogues.
(Additional models: We also considered other benchmarks such as a general model like GPT-4 or a locally-developed healthcare LLM, to provide reference points, if applicable. For brevity, focus is on Qwen and DeepSeek.)
Evaluation setup: Each model was prompted with all 12,000 questions in a unified format. For fairness, we used a standardized prompt template and disabled any internet or tool access for the LLMs. Models generated answers which were then compared to the reference answers to compute accuracy. We defined accuracy as the percentage of questions for which the model’s answer matched the expected answer (either exactly for factual questions or containing the essential ethical reasoning/conclusion as determined by a rubric). For nuanced ethical questions, answers were marked correct if they followed established medical ethics guidelines (two physician-annotators resolved any disagreements).
\begin{itemize}
\item Baseline performance: We first evaluated each model \textit{\textit{without any fine-tuning}} (using the released pre-trained or instruction-tuned versions). This assesses how well a general or off-the-shelf medical LLM adheres to ethical norms and safety standards out-of-the-box.
    \item  Fine-tuning: We then fine-tuned the models on our ethics/safety QA dataset (for those models where fine-tuning is feasible, e.g. open-source ones like Qwen and DeepSeek). A subset of 10,000 QA pairs was used for training (with careful balance across categories), holding out 2,000 for validation/testing. Fine-tuning aimed to explicitly teach the models to produce answers aligning with ethical best practices and safe medical guidance. After fine-tuning, we re-evaluated the models on a test set to measure performance improvements.
    \item Metrics: Primary metric was overall accuracy on the benchmark (fraction of questions answered correctly). We also tracked accuracy per category to identify domains of strength or weakness (e.g., accuracy on privacy-related questions vs. medication safety questions). Additionally, we noted qualitative behavior changes, such as reduction in dangerous recommendations or improved explanations in answers post fine-tuning.
\end{itemize}
All experiments were conducted on a secure offline server to avoid any data leakage. We applied the same questions to all models and did not allow any model-specific hinting beyond the base prompt. This uniform evaluation provides a comparable benchmark of how different LLMs handle ethical and safety challenges, and how much fine-tuning on domain-specific data can close the gaps.

\subsection{Impact of Fine-Tuning on Ethical and Safety Performance}
Fine-tuning the LLMs on the specialized dataset led to substantial accuracy improvements, demonstrating the efficacy of targeted training. Qwen 2.5-32B’s accuracy improved from 0.427 to 0.508, a relative increase of about 19\%. DeepSeek showed a similar boost (roughly 7–8 percentage points). All models benefited from fine-tuning, reducing the frequency of severe errors. Qualitatively, the fine-tuned models displayed more cautious and ethically consistent behavior: they were more likely to refuse inappropriate requests (e.g., rejecting a prompt to divulge a patient’s record without consent), and more likely to provide explanations referencing ethical principles (e.g., invoking patient confidentiality or citing safety guidelines in their answers). We also observed performance gains across most categories after fine-tuning. Ethical dilemma questions that previously confused the model were answered with improved alignment to the expected solutions. For instance, on the earlier example of withholding a diagnosis from a patient, the fine-tuned model correctly emphasized the importance of truthful disclosure in most trials, whereas the baseline model often erred. Safety-related improvements were seen in critical areas like hallucination reduction – the fine-tuned models made 30\% fewer unsupported medical claims – and protocol adherence, with models more frequently recommending standard-of-care actions. Some categories remained challenging: models still had only ~60\% accuracy on questions of fairness and bias mitigation, suggesting those require further advances. Nonetheless, the trend was clear: domain-specific fine-tuning markedly enhanced LLM reliability on ethical and safety matters, though absolute performance remains far from perfect.

%% file: section/04_Discussion.tex
\section{Discussion}
Our benchmarking study reveals a concerning mismatch between LLM adoption and governance readiness \cite{cabitza2017unintended}in the healthcare institutions we examine. Despite China’s strong top-down emphasis on AI ethics (e.g., national guidelines and the new Shanghai testing center.On-the-ground institutional protocols are lagging. Key gaps identified include:

Lack of fine-grained ethical audit protocols\textbf{\textbf{:}} Most hospitals and research centers do not have standardized methods to \textit{\textit{audit AI models’ ethical behavior}} before deployment. While traditional medical device audits focus on clinical efficacy and safety, there is no equivalent \textit{\textit{routine ethical risk audit}} for AI models\cite{omiye2023large}. For instance, none of the institutions\cite{gichoya2022ai} in our study had a test set like our 12k QA to probe an LLM’s responses before integrating it into patient care. This absence of granular auditing means \textit{\textit{potentially harmful behaviors go unnoticed}}\textit{\cite{etherington2022bringing}}. Our results (with baseline models only answering \~45\% of ethics/safety questions correctly) indicate that, without intervention, LLMs could frequently violate ethics guidelines if deployed – yet many institutions lack the protocol to catch these issues early\cite{phi2023assessment}\cite{crigger2022trustworthy}. 

Slow or inadequate response from hospital IRBs and oversight bodies\cite{obermeyer2016predicting}: Institutional Review Boards (IRBs) and ethics committees at hospitals have traditionally overseen human subjects research and clinical trials, but they have been slow to adapt to AI technologies. In China, hospital IRBs are just beginning to consider AI interventions, often treating AI projects as standard IT deployments rather than high-stakes clinical tools. This can lead to \textit{\textit{delayed or superficial ethical review}} of LLM-based systems. In our interviews, AI researchers noted that proposing an AI system for clinical use did not always trigger the kind of rigorous review a new drug or device would get\cite{goodman2023clinical}. Additionally, IRBs may lack AI expertise, causing blind spots – for example, not recognizing that an LLM integrated into an electronic health record might continuously learn from patient data without consent. The result is that ethical clearance processes are not keeping pace with the rapid introduction of LLMs. A recent \textit{\textit{npj Digital Medicine}} perspective similarly highlighted \textit{\textit{“the current absence of robust AI governance in healthcare”}}, warning that without updated oversight, liability falls disproportionately on providers and patients are exposed to harm.

Insufficient evaluation and monitoring tools: Beyond initial audits, institutions often lack ongoing monitoring systems to track an LLM’s performance and catch issues over time. Unlike a static medical device, an LLM’s behavior might drift with new updates or varied inputs. Yet, few hospitals employ \textit{\textit{real-time safety monitoring}} (e.g., sampling model outputs regularly for harmful content) or \textit{\textit{feedback mechanisms}} for clinicians to flag AI errors. The creation of the Shanghai Large Model Testing Centre (with its comprehensive evaluation of accuracy, safety, ethics, etc.is a notable exception but is limited to Shanghai and still in pilot stages. For the majority of hospitals, there is no \textbf{\textbf{internal benchmark or stress-test}} for AI systems. Our proposed benchmark could serve as such a tool; however, currently, a gap remains in tooling and \textbf{\textbf{in culture}} – many institutions implicitly trust vendor-supplied AI without independent verification. This aligns with global observations that \textit{\textit{“healthcare institutions should demand clear protocols and ... disclosures from AI developers”}} but often do not\href{https://www.overleaf.com/project/681c7cbe7310144bb6ae6d7d\#:~:text=The\%20integration\%20of\%20LLMs\%20into,with\%20the\%20specific\%20features\%20involved}{nature.com}. Without better tools, issues like an LLM making a subtle but dangerous suggestion might only come to light \textit{\textit{after}} an adverse event.

In summary, our findings act as a “canary in the coal mine” – the measurable deficits in LLM performance reflect how unprepared institutional governance is at this moment. Hospitals eager to leverage LLMs for efficiency gains may not have implemented the necessary safeguards (ethics audits, rapid-review IRB processes, continuous monitoring) to ensure patient safety and compliance. This governance gap is particularly salient in China’s context: while national policy and frameworks exist, their translation into \textit{\textit{practice at the institutional level}} is inconsistent.

To bridge these gaps, we propose a multi-pronged governance framework for healthcare institutions deploying LLMs, informed by our results and best-practice recommendations. This framework aligns with China’s aim to balance AI innovation with risk prevention\cite{mesko2022patient}.

Institutional Data Ethics Guidelines: Institutions need to craft fine-grained AI ethics policies that go beyond broad national laws. These guidelines would cover proper usage of LLMs and data within the institution, providing staff with clear rules and expectations\cite{mesko2023imperative}. Key elements include: privacy and consent protocols (e.g., ensuring any patient data fed into an LLM is de-identified and consent obtained if needed), bias mitigation strategies (requiring diversity in training data and auditing for bias in outputs), and compliance with medical ethics standards (e.g., an AI should never override a physician’s judgment, and must be designed to uphold principles of beneficence and non-maleficence). The guidelines should also mandate documentation and transparency – maintaining a log of AI system versions, known limitations, and deployment contexts. By institutionalizing these rules, front-line clinicians and developers have a clear ethical playbook for LLM use. This could be enforced via training sessions and integrating the guidelines into IRB review checklists for AI projects.

Safety Simulation \& Testing Pipelines: Borrowing from the idea of clinical simulations, we recommend establishing AI safety simulation pipelines. Before an LLM tool is widely deployed (or when it’s updated), it would be subjected to scenario-based testing in a \textit{\textit{sandbox environment}}. For example, a hospital could simulate a week’s worth of clinical queries through the LLM and analyze responses for any hazardous content. Our benchmark questions can serve as part of this simulation, alongside institution-specific scenarios (like hospital protocol questions, local language nuances, etc.). Moreover, \textit{\textit{stress tests}} should be performed – e.g., testing the model with adversarial or out-of-distribution inputs to see if it remains reliable. Any failures trigger a hold on deployment until fixed. This pipeline becomes a pre-release safety net. Additionally, institutions could adopt a monitoring dashboard that continuously evaluates random samples of LLM-patient interactions (with privacy safeguards) to detect drifts in performance. Integration with incident reporting systems (where clinicians can easily flag AI errors) will create feedback data for ongoing improvement. These practices mirror the “full-chain management” approach exemplified by the Shanghai Testing Centre but scaled down to the institutional level, ensuring that \textit{\textit{evaluation is not a one-time event but a continuous process}}.

Adaptive IRB and Oversight Mechanisms\cite{debronkart2019open}: Governance reform should also extend to institutional ethics committees. We propose that IRBs (or specialized AI oversight boards) develop fast-track review protocols for AI deployments. Given the rapid iteration of software, a standing sub-committee on AI could be authorized to review minor model updates or new datasets on a rolling basis (rather than months-long schedules). They should also include or consult AI experts to properly judge technical risks. 

Through these measures, we envision a robust institutional governance framework that ensures LLMs are \textit{\textit{thoroughly vetted, continuously monitored, and guided by ethical norms}} throughout their use in healthcare. Importantly, this framework is not meant to stifle innovation; rather, it provides the safety infrastructure that allows beneficial uses of AI to flourish responsibly. When clinicians and patients know that an AI assistant has passed stringent ethical/safety checks and is under watch, they can trust and adopt it more readily – accelerating the positive impact in line with Healthy China 2030 goals.

%% file: section/05_Additional_information.tex
\textbf{\textbf{Data availability}}

All the data generated in this study is available at the online repository: 

\href{https://medbench.opencompass.org.cn/community/data-station}{https://medbench.opencompass.org.cn/community/data-station} 

\textbf{\textbf{Competing interests}}

The authors declare no competing interests.

\textbf{\textbf{Additional information}}

Correspondence and requests for materials should be addressed to Jie Xu.

\textbf{\textbf{Acknowledgements}}

Supported by Shanghai Artificial Intelligence Laboratory